\documentclass[sigconf]{acmart}
\AtBeginDocument{%
  }

\setcopyright{acmlicensed}
\copyrightyear{2018}
\acmYear{2018}
\acmDOI{XXXXXXX.XXXXXXX}
\acmConference[Conference acronym 'XX]{Make sure to enter the correct
  conference title from your rights confirmation email}{June 03--05,
  2018}{Woodstock, NY}
\acmISBN{978-1-4503-XXXX-X/2018/06}

\usepackage{multirow} 
\usepackage{xcolor}
\usepackage{tcolorbox} 

\definecolor{promptheader}{RGB}{103, 168, 207} 
\definecolor{promptbody}{RGB}{248, 248, 248}
\definecolor{promptframe}{RGB}{103, 168, 207}
\begin{document}

% \title{DuEASE: An Evolving Agentic Service Engine for\\ Immersive Human-Agent Interaction}
\title{DuCCAE: A Hybrid Engine for Immersive Conversation via Collaboration, Augmentation, and Evolution}

\settopmatter{authorsperrow=4}


\author{Xin Shen}
\email{shenxin02@baidu.com}
% \affiliation{%
%   \institution{The University of Queensland}
%   \city{Brisbane}
%   \country{Australia}
% }
\affiliation{%
  \institution{Baidu Inc}
  \city{Beijing}
  \country{China}
}

\author{Zhishu Jiang}
\email{jiangzhishu@baidu.com}
\affiliation{%
  \institution{Baidu Inc}
  \city{Beijing}
  \country{China}
}

\author{Jiaye Yang}
\email{yangjiaye01@baidu.com}
\affiliation{%
  \institution{Baidu Inc}
  \city{Beijing}
  \country{China}
}

\author{Haibo Liu}
\email{liuhaibo05@baidu.com}
\affiliation{%
  \institution{Baidu Inc}
  \city{Beijing}
  \country{China}
}

\author{Yichen Wan}
\email{wanyichen@baidu.com}
\affiliation{%
  \institution{Baidu Inc}
  \city{Beijing}
  \country{China}
}

\author{Jiarui Zhang}
\email{zhangjiarui05@baidu.com}
\affiliation{%
  \institution{Baidu Inc}
  \city{Beijing}
  \country{China}
}

\author{Tingzhi Dai}
\email{daitingzhi@baidu.com}
\affiliation{%
  \institution{Baidu Inc}
  \city{Beijing}
  \country{China}
}

\author{Luodong Xu}
\email{xuluodong@baidu.com}
\affiliation{%
  \institution{Baidu Inc}
  \city{Beijing}
  \country{China}
}

\author{Shuchen Wu}
\email{wushuchen@baidu.com}
\affiliation{%
  \institution{Baidu Inc}
  \city{Beijing}
  \country{China}
}

\author{Guanqiang Qi}
\email{qiguanqiang@baidu.com}
\affiliation{%
  \institution{Baidu Inc}
  \city{Beijing}
  \country{China}
}

\author{Chenxi Miao}
\email{miaochenxi@baidu.com}
\affiliation{%
  \institution{Baidu Inc}
  \city{Beijing}
  \country{China}
}

\author{Jiahui Liang}
\email{liangjiahui03@baidu.com}
\affiliation{%
  \institution{Baidu Inc}
  \city{Beijing}
  \country{China}
}

\author{Weikang Li}
\authornote{Corresponding authors: Weikang Li and Jizhou Huang.}
\email{wavejkd@pku.edu.cn}
\affiliation{%
  \institution{Baidu Inc}
  \city{Beijing}
  \country{China}
}

\author{Yang Li}
\email{liyang164@baidu.com}
\affiliation{%
  \institution{Baidu Inc}
  \city{Beijing}
  \country{China}
}

\author{Deguo Xia}
\email{xiadeguo@baidu.com}
% \affiliation{%
%   \city{Beijing}
%   \country{China}
% }
\affiliation{%
  \institution{Baidu Inc}
  \city{Beijing}
  \country{China}
}

\author{Jizhou Huang\footnotemark[1]}
\authornote{Project lead: Jizhou Huang.}
\email{huangjizhou01@baidu.com}
\affiliation{%
  \institution{Baidu Inc}
  \city{Beijing}
  \country{China}
}

\renewcommand{\shortauthors}{Xin Shen et al.}
\renewcommand{\shorttitle}{Baidu DuCCAE System}

%%%%%%%%  Paper Begin %%%%%%%%%%%%
\begin{abstract}
Immersive conversational systems in production face a persistent trade-off between responsiveness and long-horizon task capability.
Real-time interaction is achievable for lightweight turns, but requests involving planning and tool invocation (\emph{e.g.}, search and media generation) produce heavy-tail execution latency that degrades turn-taking, persona consistency, and user trust.
To address this challenge, we propose \textbf{\textit{DuCCAE}} (\underline{\textbf{C}}onversation while \underline{\textbf{C}}ollaboration with \underline{\textbf{A}}ugmentation and \underline{\textbf{E}}volution), a hybrid engine for immersive conversation deployed within \textbf{Baidu Search}, serving millions of users.
\textit{DuCCAE} decouples real-time response generation from asynchronous agentic execution and synchronizes them via a shared state that maintains session context and execution traces, enabling asynchronous results to be integrated back into the ongoing dialogue.
The system orchestrates five subsystems—\textit{Info, Conversation, Collaboration, Augmentation, and Evolution}—to support multi-agent collaboration and continuous improvement.
We evaluate \textit{DuCCAE} through a comprehensive framework that combines offline benchmarking on the \textbf{Du-Interact} dataset and large-scale production evaluation within \textbf{Baidu Search}.
Experimental results demonstrate that \textit{DuCCAE} outperforms strong baselines in agentic execution reliability and dialogue quality while reducing latency to fit strict real-time budgets.
Crucially, deployment metrics since June 2025 confirm substantial real-world effectiveness, evidenced by a tripling of Day-7 user retention to 34.2\% and a surge in the complex task completion rate to 65.2\%.
Our hybrid architecture successfully preserves conversational continuity while enabling reliable agentic execution, offering practical guidelines for deploying scalable agentic systems in industrial settings.
\end{abstract}



\maketitle

\begin{figure*}[t]
  \centering
  \includegraphics[width=0.98\linewidth]{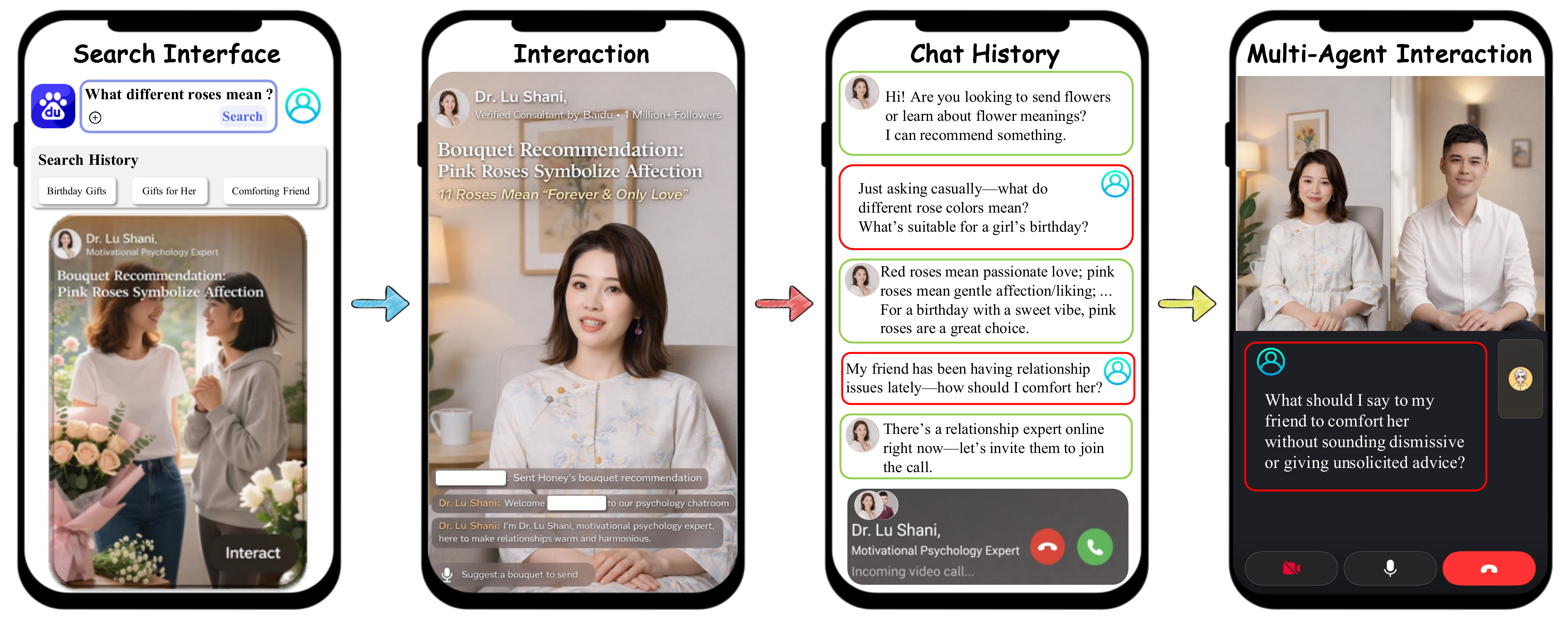}
\caption{\textit{\textbf{DuCCAE}} interface and interaction flow in production. Starting from a \textbf{Baidu} Search entry, the system maintains persona-consistent chat responses and supports escalation to real-time calling; complex requests trigger asynchronous collaboration and tool augmentation while preserving conversational continuity.
}
  \label{fig:interface}
\vspace{-1em}
\end{figure*}

\section{Introduction}
\label{sec:intro}
Immersive conversational interfaces are rapidly becoming the primary entry point for diverse production domains, ranging from search engines to customer support and intelligent assistants~\cite{DBLP:journals/corr/abs-2509-21574,DBLP:journals/corr/abs-2511-12662,DBLP:journals/tois/MoMZQCCLZDN25,DBLP:journals/inffus/SapkotaRK26,yang-etal-2025-spoken}. 
Users in these environments demand a seamless blend of low-latency response and complex task execution~\cite{DBLP:journals/corr/abs-2410-21620,DBLP:journals/corr/abs-2402-02716,DBLP:journals/corr/abs-2506-23826}. 
Such interactions frequently involve intricate reasoning and tool usage grounded in multimodal signals (\emph{e.g.}, speech, video, and text)~\cite{park2024let,DBLP:conf/cvpr/0001DD0ZCL025,DBLP:journals/corr/abs-2406-15177}. 
Processing these inputs under strict real-time budgets introduces a critical bottleneck, where the heavy-tail latency inherent in agentic planning breaks conversational flow and destabilizes persona consistency~\cite{DBLP:journals/corr/abs-2410-21620,DBLP:journals/corr/abs-2402-02716,DBLP:journals/dase/XuHGS25}.

While recent agent and multi-agent paradigms have significantly improved capability via planning, routing, and tool invocation, they effectively exacerbate this latency bottleneck~\cite{DBLP:journals/corr/abs-2402-02716,DBLP:journals/corr/abs-2308-08155,DBLP:journals/corr/abs-2403-16971}. 
Long-horizon execution and tool latency are notoriously difficult to bound in production environments~\cite{DBLP:journals/corr/abs-2410-21620,DBLP:journals/corr/abs-2510-00481}. 
Consequently, ``capability-first'' pipelines often achieve strong task performance yet degrade interaction quality due to silence, interruptions, and inconsistent persona, whereas ``responsiveness-first'' conversational setups preserve fluency but struggle with reliable tool use and long-horizon task completion~\cite{DBLP:conf/acl/MaharanaLTBBF24,DBLP:journals/corr/abs-2406-15177,DBLP:journals/corr/abs-2404-11584}.
Bridging responsiveness and capability in a single production system therefore requires a serving architecture that can keep the dialogue responsive while executing complex tasks asynchronously, and can integrate intermediate and final outcomes back into the same conversational context~\cite{DBLP:journals/corr/abs-2502-17419}.

To address this challenge, we present \textbf{\textit{DuCCAE}}, a hybrid engine for immersive conversation developed and deployed at \textbf{Baidu}~\cite{ernie2025technicalreport,ernie2026technicalreport}. 
As illustrated in Figure~\ref{fig:interface}, \textit{DuCCAE} instantiates this paradigm within the \textit{Baidu Search} interface via an interactive digital human~\cite{DBLP:conf/ACMdis/RashikJKSM24,park2024let}. 
Crucially, the system utilizes a decoupled architecture that separates real-time video-based rendering from asynchronous agentic execution~\cite{yu2025llia,DBLP:journals/cacm/FabianoGLMMPRSV25,DBLP:journals/corr/abs-2510-09608}. 
To achieve this, \textit{DuCCAE} orchestrates five tightly integrated subsystems---Info, Conversation, Collaboration, Augmentation, and Evolution---to manage the complex interplay between multimodal perception, real-time response, and long-horizon tool use~\cite{DBLP:conf/acl/MaharanaLTBBF24}.
This design allows the digital human to sustain immediate, persona-consistent responsiveness during simple turns, while long-horizon tasks are offloaded to a background multi-agent collaboration layer~\cite{DBLP:conf/acl/MaharanaLTBBF24,DBLP:journals/corr/abs-2510-11967}. 
A unified shared state bridges these two paths by retaining session context and execution traces, enabling tool outputs and agent deliverables to be integrated back into the live video interaction seamlessly without breaking immersion~\cite{DBLP:conf/iclr/YaoZYDSN023,DBLP:journals/corr/abs-2410-08328,DBLP:journals/corr/abs-2503-23278}. 
Moreover, \textit{DuCCAE} operates as an evolving service engine, where real interactions continuously drive automated evaluation and post-training updates~\cite{evo1,evo2}.

\begin{figure*}[t]
  \centering
  \includegraphics[width=0.98\linewidth]{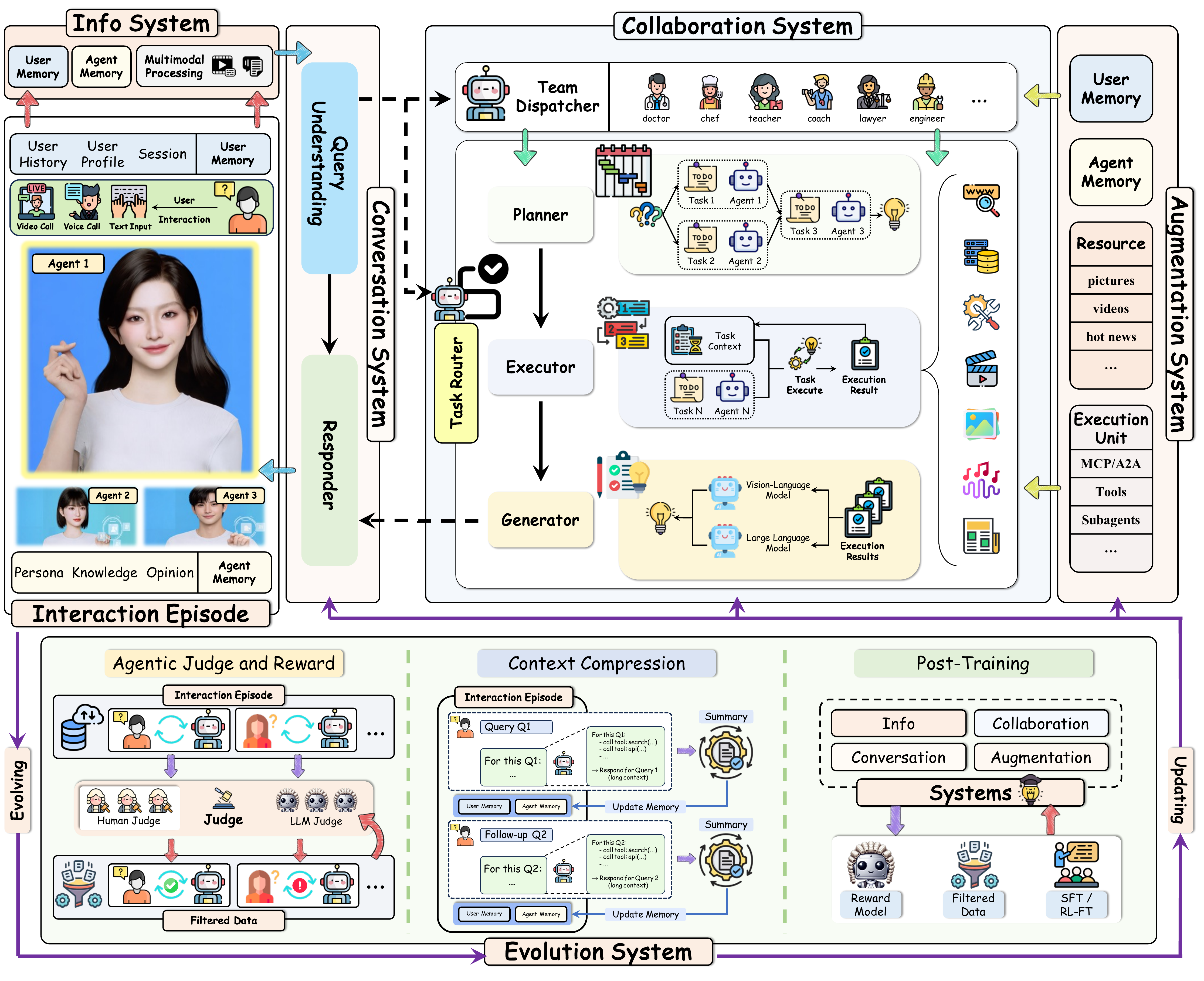}
  \caption{\textbf{\textit{DuCCAE}}: an evolving agentic service engine for immersive conversational interaction. 
  \textit{Info System} converts multimodal signals into policy-aware context and manages memory. 
  \textit{Conversation System} acts as a low-latency gatekeeper for intent routing and renders persona-consistent responses. 
  \textit{Collaboration System} supports multi-agent execution for long-horizon tasks via planning and tool use. 
  \textit{Augmentation System} empowers agents with external tools, retrieval resources, and execution protocols. 
  \textit{Evolution System} drives continuous improvement through episode-based judging and post-training.}
  \label{fig:system-arch}
\vspace{-0.5em}
\end{figure*}

To validate the effectiveness of this hybrid engine, we evaluate \textit{DuCCAE} in a production setting using both online experiments and post-launch metrics.
The key contributions to both the research and industrial communities are as follows: 

\begin{itemize}
    \item We propose a decoupled framework that resolves the trade-off between responsiveness and capability. 
    By synchronizing real-time and asynchronous streams, we ensure sub-second latency while supporting complex tool use.
    
    \item We demonstrate that our evolved model outperforms zero-shot baselines with significantly larger parameters in both dispatch precision and task success rate.
    
    \item Deployed to millions of users since June 2025, \textit{DuCCAE} has driven substantial gains, achieving a \textbf{3$\times$ increase in Day-7 User Retention (34.2\%)} and a \textbf{Complex Task Completion Rate of 65.2\%}.
\end{itemize}

\section{System Overview}
\label{sec:system}

Figure~\ref{fig:system-arch} illustrates the architecture of \textbf{\textit{DuCCAE}}, formulated as a hybrid orchestration framework designed to bridge the gap between real-time responsiveness and agentic capability in immersive settings~\cite{adimulam2026orchestration,DBLP:journals/corr/abs-2410-08328,DBLP:journals/corr/abs-2308-08155}. 
Unlike monolithic pipelines that block on tool execution, \textit{DuCCAE} implements a Latency-Decoupled Architecture~\cite{DBLP:conf/cui/MaslychKLHGPPME25}. 
The core design philosophy is to separate the \textit{interaction loop} (which demands strictly bounded latency for immersion) from the \textit{execution loop} (which involves heavy-tail reasoning for complex tasks), utilizing a Unified Shared State as the synchronization barrier~\cite{zhang2025agentorchestra,DBLP:journals/corr/abs-2505-12501}. 
This architecture instantiates the proposed decoupled paradigm through five tightly integrated subsystems: \textbf{Info}, \textbf{Conversation}, \textbf{Collaboration}, \textbf{Augmentation}, and \textbf{Evolution}.

\subsection{Dual-Track Dataflow and Synchronization}
\label{sec:dataflow}
The runtime behavior of \textit{DuCCAE} is governed by a \textbf{Dual-Track Execution Mechanism}~\cite{DBLP:journals/corr/abs-2510-08731,DBLP:journals/corr/abs-2510-00202} that dynamically routes multimodal streams based on computational complexity, as illustrated in Figure~\ref{fig:execution_flow}.
Given an interaction episode, the system employs a semantic routing strategy to direct the execution flow through two coordinated paths: a \textit{Fast Track} for immediate response and a \textit{Slow Track} for asynchronous reasoning, which are eventually reunited via \textit{Event-Driven Synchronization}. 
The \textit{Query Understanding} module (detailed in Section~\ref{sec:conversation}) executes this critical routing selection by evaluating the intent complexity~\cite{varangot2025doing}.

\textit{The Fast Track for Real-Time Interaction.} 
The first path is the Fast Track, dedicated to maintaining conversational immersion by bypassing heavy reasoning modules. 
It triggers the \textit{Conversation System} to directly resolve lightweight queries by leveraging pre-loaded agent knowledge and user context~\cite{DBLP:journals/corr/abs-2412-15266}. 
Such a streamlined process guarantees that persona-consistent feedback, including non-verbal cues or memory-grounded replies, is generated within a strict Time-to-First-Token (TTFT) budget of under 500ms~\cite{iliakopoulou2025chameleon}.

\textit{The Slow Track for Asynchronous Reasoning.} 
The second path is the Slow Track, which handles complex intents by offloading planning and tool invocation to the \textit{Collaboration System}~\cite{DBLP:journals/corr/abs-2311-17541,DBLP:conf/icml/ErdoganL0MFAKG25}.
Operating in the background, it executes long-horizon tasks without blocking the real-time interaction loop, thereby allowing the digital human~\cite{guan2024talk,guan2024resyncer,yang2024showmaker} to maintain engagement while complex work proceeds asynchronously.

\textit{Event-Driven Synchronization.} 
These two tracks are synchronized via an Event-Driven Integration mechanism~\cite{zhang2025agentorchestra,DBLP:journals/corr/abs-2505-12501}. 
When the Slow Track produces intermediate artifacts or final results, it emits a state-update event. 
Upon capturing these signals, the \textit{Shared State} prompts the real-time path to seamlessly incorporate the new information into the ongoing video stream. 
Consequently, the user perceives a single and coherent dialogue flow rather than disjoint system outputs.

\subsection{Subsystem Overview}
\label{sec:subsystem_overview}

Having established the dual-track orchestration mechanism, we now detail the design rationale and specific responsibilities of each constituent subsystem, following the information flow from perception to evolution.

\subsubsection{Info System: Multimodal Perception and State Construction} \label{sec:info}
The \textbf{Info System} serves as the foundational perception layer, tasked with converting raw, unstructured interaction signals into a structured and policy-aware state space. 
To balance the critical trade-off between perception depth and real-time latency, the architecture aligns all modalities into a unified textual representation. 
Such a design enables the system to leverage powerful Large Language Models (LLMs)~\cite{ernie2025technicalreport,ernie2026technicalreport,DBLP:journals/corr/abs-2502-13923} for reasoning without incurring the computational overhead typically associated with processing high-dimensional video streams end-to-end.

The system processes heterogeneous inputs through specialized pipelines to extract textual semantic cues. 
For audio streams, a streaming Automatic Speech Recognition (ASR)~\cite{amodei2016deep,hannun2014deep} engine converts voice input into text with millisecond-level latency. 
Concurrently, handling video streams requires a distinct strategy to mitigate the heavy-tail latency inherent in large Vision-Language Models~\cite{ernie2025technicalreport,ernie2026technicalreport}. 
% Captioning progress has time latency, could use the perivous summarised visual info for current dialoge. 
The system therefore employs a lightweight VLM dedicated to efficient semantic extraction. 
Instead of dense frame-by-frame inference, this module generates concise captions describing key visual elements, including user actions, detected objects, and environmental tags. 
These structured text descriptions are then aligned with the ASR output to form a cohesive multimodal context.

\begin{figure*}[t]
  \centering
  \includegraphics[width=0.98\linewidth]{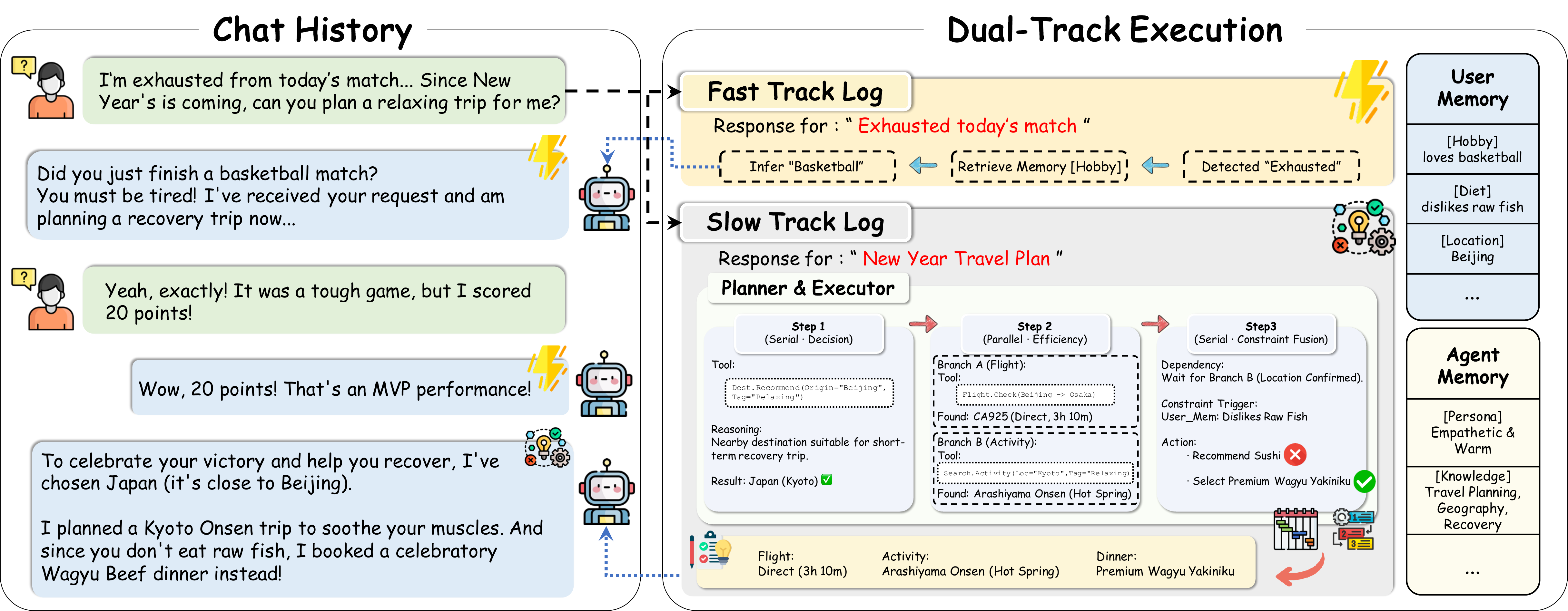}
\vspace{-0.5em}
\caption{\textbf{Runtime Execution Dataflow of a Mixed-Intent Interaction Episode.} 
Processing the input ``Exhausted... Plan a trip'', the \textbf{Fast Track} (top) utilizes User Memory (``Hobby: Basketball'') for an immediate empathetic inquiry. 
Simultaneously, the \textbf{Slow Track} (bottom) decomposes the travel request, demonstrating: 
(1) \textbf{Parallel Execution} of independent flight and activity searches to optimize latency; and 
(2) \textbf{Constraint Fusion}, which filters dining options via the memory constraint (``Dislikes Raw Fish'') to select the compatible ``Wagyu Beef''.}
  \label{fig:execution_flow}
\vspace{-1em}
\end{figure*}

Beyond perception, the Info System orchestrates a dual-memory mechanism (as depicted in Figure~\ref{fig:system-arch}) that strictly decouples user personalization from the agent execution context~\cite{DBLP:journals/corr/abs-2412-15266,DBLP:journals/corr/abs-2502-06975}.
\textbf{User Memory} aggregates persistent signals to construct a stable user model, comprising static attributes in the \textit{User Profile}, retrieval-augmented logs in the \textit{User History}, and the immediate turn-taking context in the \textit{Session State}. 
Fusion of these signals ensures that responses remain grounded in the specific persona and constraints of the user. 
Distinct from these personal attributes, \textbf{Agent Memory} functions as the unified cognitive storage for the digital human. 
As illustrated in Figure~\ref{fig:execution_flow}, it houses the agent's \textbf{static profiles} (Persona and Knowledge Base) to guide consistent role-play, while primarily serving as the \textbf{working memory} to record the System-of-Thought (SoT) trajectory, which includes the current plan generated by the Collaboration System and intermediate tool traces~\cite{DBLP:journals/corr/abs-2311-17541,DBLP:journals/corr/abs-2412-15266}.
Isolating the execution state allows the agent to manage complex, ephemeral task data without polluting the long-term user profile.

Following this stratification, the \textit{Responder} module orchestrates the output delivery. 
For immediate queries, it directly synthesizes the final response. 
Crucially, for complex requests requiring long-horizon reasoning, the Responder employs a ``bridge-and-callback'' strategy. 
Upon routing the request, it immediately generates a semantic bridging response—such as the acknowledgement ``\textit{I've received your request...}'' shown in the Fast Track of Figure~\ref{fig:execution_flow}—to prevent ``dead air'' and maintain conversational momentum~\cite{DBLP:conf/cui/MaslychKLHGPPME25}. 
Once the asynchronous workflow completes, the results are seamlessly reintegrated into the ongoing dialogue stream.

\subsubsection{Conversation System: Latency-Aware Interaction and Routing} \label{sec:conversation}

The \textbf{Conversation System} operates as the real-time interaction hub, strictly bounded by a latency budget to ensure immersive responsiveness. 
Functioning as the gatekeeper for the dual-track mechanism, it executes two critical processes: intent stratification via \textit{Query Understanding} and stream-based rendering via the \textit{Responder}.

The \textit{\textbf{Query Understanding}} module acts as the decision engine, parsing the normalized Request Object to map user inputs onto a three-tiered complexity hierarchy~\cite{varangot2025doing,DBLP:journals/corr/abs-2510-08731}. 
This classification explicitly dictates the subsequent execution path:

\begin{enumerate} 

\item \textbf{Lightweight Queries (Tier-1):} Interactions that can be resolved immediately using pre-loaded context and general knowledge (\emph{e.g.}, chit-chat, greetings, or memory-based QA). 
These are routed exclusively to the Fast Track for sub-second response.

\item \textbf{Deterministic Tool Intents (Tier-2):} Requests where the user explicitly seeks external information requiring discrete API calls (\emph{e.g.}, querying weather, stock prices, or calendar entries). These trigger specific tools within the Augmentation System while maintaining a simple execution flow.

\item \textbf{Complex Domain Requests (Tier-3):} High-order tasks necessitating intricate reasoning, multi-step planning, or professional expertise (\emph{e.g.}, medical consultation, travel itinerary planning, or legal advice). These leverage the \textit{Team Dispatcher} to recruit specialized agents within the Collaboration System for asynchronous collaboration.
\end{enumerate}

Following this stratification, the \textit{\textbf{Responder}} module generates the appropriate user-facing output. For Tier-1 queries, it directly synthesizes a final response grounded in the active persona. Conversely, for Tier-2 and Tier-3 requests that trigger asynchronous workflows, the Responder generates a semantic bridging response, such as a clarification question or a verbal acknowledgement (``I will check the latest reports for you...''), to maintain the conversational floor and mask the execution latency.

\subsubsection{Collaboration System: Asynchronous Multi-Agent Execution} \label{sec:collaboration}

The \textbf{Collaboration System} serves as the execution engine for the ``Slow Track'', handling Tier-2 and Tier-3 requests that exceed the capabilities of immediate retrieval. 
To ensure reliability and efficiency under heavy-tail latency, the system implements a standardized \textit{Plan--Execute--Generate} workflow~\cite{DBLP:journals/corr/abs-2311-17541,DBLP:conf/icml/ErdoganL0MFAKG25}.

The process initiates with the \textit{\textbf{Team Dispatcher}}, which functions as a semantic routing layer. 
The system maintains a library of specialized agents (\emph{e.g.}, medical experts, legal advisors, or data analysts), each characterized by a specific model profile and knowledge base. 
Upon receiving a complex request, the Dispatcher employs a semantic matching mechanism to align the query's intent embedding with the most suitable agent profile. 
This dynamic recruitment ensures that specialized domain knowledge is applied precisely to the corresponding problem space.

Once a specialized agent is instantiated, the workflow enters the \textbf{\textit{Planner}} phase, where the primary objective is latency optimization through structural decomposition. 
The Planner analyzes the request to construct a task dependency graph, distinguishing between serial tasks and parallel tasks~\cite{patelsix,DBLP:journals/corr/abs-2505-12501}. 
As visually demonstrated in the ``Slow Track Log'' of Figure~\ref{fig:execution_flow}, the Planner schedules the flight check and activity search concurrently (Step 2) to optimize efficiency, while strictly enforcing sequential dependencies for dining reservations (Step 3).
By identifying non-blocking sub-tasks, the Planner enables concurrent tool invocation, significantly accelerating the overall execution timeline.

Following the plan, the \textbf{\textit{Executor}} carries out the concrete actions. 
It adheres to an iterative state update protocol: as each step is executed, the intermediate results (\emph{e.g.}, API returns or sub-agent replies) are written back into the shared Task Context. 
This mechanism is critical for serial tasks, as it ensures that subsequent steps have access to the data generated by their predecessors, maintaining logical continuity throughout the problem-solving chain.

The workflow concludes with the \textbf{\textit{Generator}}, which synthesizes the aggregated execution traces into a coherent user-facing deliverable. 
This module is model-agnostic and modality-aware: it selects a Large Language Model (LLM)~\cite{ernie2025technicalreport,ernie2026technicalreport,DBLP:journals/corr/abs-2412-15115} for textual summaries or a Vision-Language Model (VLM)~\cite{ernie2025technicalreport,ernie2026technicalreport,DBLP:journals/corr/abs-2502-13923} when visual synthesis is required. 
The final precise output is then synchronized back to the shared state, ready to be presented by the digital human~\cite{guan2024talk,guan2024resyncer,yang2024showmaker}.

\begin{table*}[t]
  \centering
  \caption{Statistics of \textbf{Du-Interact-Evo} (Training) and \textbf{Du-Interact} (Testing) across Evolutionary Stages.}
  \vspace{-0.5em}
  \label{tab:dataset_stats}
  \begin{tabular}{llccc}
    \toprule
    \textbf{System Ver.} & \textbf{Evolution Stage} & \textbf{Training Source} & \textbf{Dataset Size} & \textbf{Method} \\
    \midrule
    DuCCAE-V1 & Cold Start & N/A & 0 & Zero-shot Prompting \\
    DuCCAE-V2 & Evolution I & \textbf{Du-Interact-Evo-V1} & 15,000 & SFT (on V1 logs) \\
    DuCCAE-V3 & Evolution II & \textbf{Du-Interact-Evo-V2} & 50,000 & SFT + RL (on V2 logs) \\
    \midrule
    \textit{Benchmark} & \textit{Evaluation} & \textit{\textbf{Du-Interact}} & \textit{5,000} & \textit{Fixed Golden Set} \\
    \bottomrule
  \end{tabular}
\vspace{-1em}
\end{table*}

\subsubsection{Augmentation System: Retrieval, Protocols, and Tool Ecosystem} 
\label{sec:augmentation}

The \textbf{Augmentation System} extends the native capabilities of the core models by bridging them with external knowledge bases and executable environments.
Rather than a static repository, it operates as a dynamic service layer that grounds model generation in reality and standardizes complex tool interactions.

To ensure response fidelity and accelerate inference, the system implements a Retrieval-Augmented Generation (RAG)~\cite{wang2024searching,chen2025pairs,chen2025cmrag} mechanism over a heterogeneous resource layer. 
This layer aggregates proprietary internal databases, real-time ``hot news'', and the structured histories from both User and Agent Memory. 
By retrieving relevant context prior to generation, the system achieves two critical objectives: \textit{Consistency} and \textit{Efficiency}. 
Retrieving historical memory ensures that the agent's behavior aligns strictly with the user's persona and past interactions, while retrieving structured knowledge allows for faster, high-confidence responses without hallucinations. 
This mechanism not only grounds the agent in verified data but also injects diversity into the conversation by surfacing rich, domain-specific content.

To manage the complexity of diverse extensions, the system abstracts all external interactions through a Unified Execution Unit interface, adhering to strict industry protocols. 
Specifically, we implement the Model Context Protocol (MCP) to standardize how context is passed between the LLM and external tools, ensuring the model accurately understands tool states~\cite{DBLP:journals/corr/abs-2503-23278}. 
Furthermore, for multi-agent handshakes, we utilize an Agent-to-Agent (A2A) protocol that defines clear contracts for task delegation and result return. 
This standardization allows different specialized agents to collaborate seamlessly without format mismatches, treating tools and sub-agents as interchangeable functional blocks.

The ecosystem supports a wide array of capabilities, ranging from information retrieval (\emph{e.g.}, Search, Document Parsing) to creative media synthesis (\emph{e.g.}, Image and Music Generation). 
Crucially, the deployment of Sub-Agents is governed by a Robust Orchestration Engine. 
When executing tasks—particularly during the parallel processing paths defined by the Planner—this engine enforces strict concurrency controls and validity checks. It prevents execution conflicts and manages failure states, ensuring that a stall in one sub-agent does not deadlock the entire workflow. 
This guarantees that complex, multi-tool chains are executed atomically and reliably.

\subsubsection{Evolution System: Data-Driven Feedback and Adaptive Compression} 
\label{sec:evolution}

The \textbf{Evolution System} functions as the engine for continuous improvement, closing the loop between deployment and development~\cite{evo1,evo2}. 
Instead of static updates, it employs a periodic \textit{Data Flywheel} mechanism that processes interaction episodes to refine both the system's efficiency and its intelligence.

To construct high-quality training datasets, raw interaction logs first undergo Automated Assessment~\cite{li2025generation,Gu2024AASO}.
Model-based judges evaluate episodes based on three pragmatic criteria: 
(1) Next-Turn Engagement, measuring whether the system's response successfully elicited a meaningful follow-up from the user (a key proxy for immersion); 
(2) Instruction Compliance, checking if tool invocations logically matched the user's intent; 
and (3) Sentiment Alignment, analyzing the emotional valence of the user's reaction. 
This automated phase produces a large-scale ``Silver Dataset'' for Supervised Fine-Tuning (SFT). 
Subsequently, a subset is sampled for Human Verification, yielding a high-precision ``Gold Dataset'' reserved for training Reward Models and Reinforcement Learning (RL) benchmarks.

Our system implements a two-stage \textit{Compression Strategy} to manage the life-cycle of information, preventing context overflow while retaining critical signals.

\begin{itemize} 
\item \textbf{Runtime Context Folding:} 
To mitigate the heavy token overhead caused by verbose tool outputs, we adopt the \textbf{Context-Folding} mechanism inspired by \cite{DBLP:journals/corr/abs-2510-00615,DBLP:journals/corr/abs-2510-11967}. 
During complex execution chains (in the Collaboration System), the agent isolates intermediate steps—such as error traces or massive JSON payloads—into temporary branches. 
Upon task completion, these branches are ``folded'' into concise semantic summaries before being merged back into the main context. 
This drastically reduces context occupancy and latency for subsequent turns.

\item \textbf{Episodic Memory Distillation:} 
At the session level, raw conversation logs are too noisy for long-term storage. 
The system applies an abstraction policy to distill completed sessions into structured Knowledge Nuggets~\cite{DBLP:journals/corr/abs-2412-15266,DBLP:journals/corr/abs-2502-06975} (\emph{e.g.}, ``User prefers visual data over text'' or ``User is a vegetarian''). 
These distilled artifacts are then committed to the long-term \textit{User Memory} and \textit{Agent Memory}, ensuring that the memory module grows in density rather than just volume.
\end{itemize}

Finally, the processed assets drive a \textit{Modular Post-Training} phase. 
The filtered ``Silver'' and ``Gold'' datasets are stratified to target specific subsystems, such as conversational pairs enrich the \textit{Conversation System's} persona adaptability, while folded execution traces refine the \textit{Collaboration System's} planning capabilities. 
By iteratively retraining on these filtered experiences, DuCCAE evolves from a static system into an adaptive agent that aligns increasingly closer to user needs over time.

\begin{table*}[t]
  \centering
  \caption{Main Results on the \textbf{Du-Interact} Golden Benchmark. 
  % We compare \textit{DuCCAE} (V1-V3) against open-source baselines ranging from 3B to 70B parameters. 
  }
  \vspace{-0.5em}
  \label{tab:main_results}
  \resizebox{0.95\linewidth}{!}{
  \begin{tabular}{ll|cc|ccc|c}
    \toprule
    \multirow{2}{*}{\textbf{Backbone Model}} & \multirow{2}{*}{\textbf{Setup}} & \multicolumn{2}{c|}{\textbf{Task Execution}} & \multicolumn{3}{c|}{\textbf{Dialogue Quality}} & \textbf{Efficiency} \\
    & & \textbf{Dispatch ($P_{disp}$)} & \textbf{Success Rate (SR)} & \textbf{Fidelity (\%)} & \textbf{Persona (1-5)} & \textbf{Empathy (1-5)} & \textbf{Avg. Latency} \\
    \midrule
    \textit{Edge Class (3B)} & & & & & & & \\
    Qwen2.5-3B-Instruct & Zero-shot & 52.4\% & 28.5\% & 45.2\% & 2.5 & 2.4 & \textbf{1,250 ms} \\
    Llama-3.2-3B-Instruct & Zero-shot & 51.8\% & 26.2\% & 44.1\% & 2.4 & 2.3 & 1,280 ms \\
    \midrule
    \textit{Server Class (7B-11B)} & & & & & & & \\
    Qwen2.5-7B-Instruct & Zero-shot & 61.5\% & 37.2\% & 54.8\% & 2.8 & 2.8 & 1,480 ms \\
    Llama-3.2-11B-Instruct & Zero-shot & 58.2\% & 35.8\% & 53.5\% & 2.7 & 2.7 & 1,520 ms \\
    \midrule
    \textit{Large Scale Class (30B-70B)} & & & & & & & \\
    Qwen2.5-32B-Instruct & Zero-shot & 72.1\% & 55.4\% & 62.2\% & 3.2 & 3.2 & 3,250 ms \\
    Llama-3.3-70B-Instruct & Zero-shot & 75.5\% & 61.1\% & 64.4\% & 3.4 & 3.3 & 5,800 ms \\
    \midrule
    \textit{\textbf{DuCCAE (Ours)}} & & & & & & & \\
    \textit{DuCCAE}-V1 & \textbf{Cold Start} & 68.8\% & 50.5\% & 58.4\% & 3.5 & 3.5 & 1,850 ms \\
    \textit{DuCCAE}-V2 & \textbf{Evolution I} & 76.2\% & 63.8\% & 66.5\% & 3.9 & 4.0 & 1,920 ms \\
    \textit{DuCCAE}-V3 & \textbf{Evolution II} & \textbf{82.5\%} & \textbf{72.4\%} & \textbf{71.1\%} & \textbf{4.1} & \textbf{4.3} & 1,880 ms \\
    \bottomrule
  \end{tabular}
  }
\vspace{-1em}
\end{table*}

\section{Experiments}
\label{sec:experiments}

To rigorously evaluate the effectiveness of \textit{DuCCAE} in a realistic industrial setting, we conduct comprehensive evaluations comparing our evolved system against a wide spectrum of state-of-the-art open-source baselines.
All experiments are conducted on \textbf{Du-Interact}, a standardized human-curated benchmark derived from the production environment of \textbf{Baidu Search}.

\subsection{Datasets and Protocols}

Our data strategy distinguishes between the dynamic training stream and the static evaluation benchmark to ensure rigorous measurement of evolutionary progress.

\subsubsection{Du-Interact-Evo: The Evolutionary Training Stream}
The training data for \textit{DuCCAE} is not static; it is generated dynamically via the proposed Evolution System. We term this continuously growing dataset \textbf{Du-Interact-Evo}. As shown in Table~\ref{tab:dataset_stats}, the system underwent three evolutionary stages:

\noindent\textbf{Stage I (Cold Start):} At the inception (V1), no domain-specific training data was available. The system relied on zero-shot prompting of the foundation model using expert-defined rules.

\noindent\textbf{Stage II (Evolution I):} We collected 3 months of interaction logs from the V1 system. 
These logs were filtered by the ``Data Flywheel'' pipeline (Automated Judging + Human Verification) to construct \textbf{Du-Interact-Evo-V1}, a high-quality Supervised Fine-Tuning (SFT) dataset containing 15,000 episodes.

\noindent\textbf{Stage III (Evolution II):} Building on V2, we expanded the collection to 50K episodes (\textbf{Evo-V2}). 
This stage introduced Reinforcement Learning (RL) signals, prioritizing sessions where users successfully completed complex tasks with high sentiment scores.

\subsubsection{Du-Interact: The Golden Test Benchmark}
To monitor the system's trajectory fairly, we construct \textbf{Du-Interact}, a fixed Golden Test Set containing 5,000 multi-turn interaction sessions sampled strictly from the V2 phase hard-negatives.
Crucially, each session in \textbf{Du-Interact} is annotated by human experts with Ground Truth (GT) labels for: (1) correct intent routing logic, (2) optimal tool selection and parameter extraction, and (3) ideal response content. 
This ensures that V1, V2, and V3 are evaluated against the same high standard.

\subsection{Baselines and Setup}

\subsubsection{System Architecture}
Our core system utilizes \textbf{ERNIE-4.5-21B-A3B}~\cite{ernie2025technicalreport} as the central reasoning backbone (Slow Track) and \textbf{ERNIE-4.5-VL-28B-A3B}~\cite{ernie2025technicalreport} for visual perception.
For fair comparison, all open-source baselines replace only the reasoning backbone while retaining the same visual front-end and the identical tool ecosystem~\cite{liu2023improvedllava,liu2023llava}.

\subsubsection{Baseline Models}

We categorize baselines into three groups to analyze the impact of model scale comprehensively:

\noindent\textit{Edge Class (3B):} We select Qwen2.5-3B-Instruct~\cite{DBLP:journals/corr/abs-2412-15115} and Llama-3.2-3B-Instruct~\cite{liu2023llava}. 
These models represent the ultra-low latency tier suitable for on-device deployment.

\noindent\textit{Server Class (7B-11B):} We select Qwen2.5-7B-Instruct~\cite{DBLP:journals/corr/abs-2412-15115} and Llama-3.2-11B-Instruct~\cite{liu2023llava}. 
These are the industry standards for cost-effective server-side inference.

\noindent\textit{Large Scale Class (30B-70B):} We select Qwen2.5-32B-Instruct~\cite{DBLP:journals/corr/abs-2412-15115} and Llama-3.3-70B-Instruct~\cite{liu2023llava}. 
These serve as powerful ``upper bound'' references for zero-shot reasoning capabilities.

\begin{figure*}[t]
  \centering
  \includegraphics[width=1\linewidth]{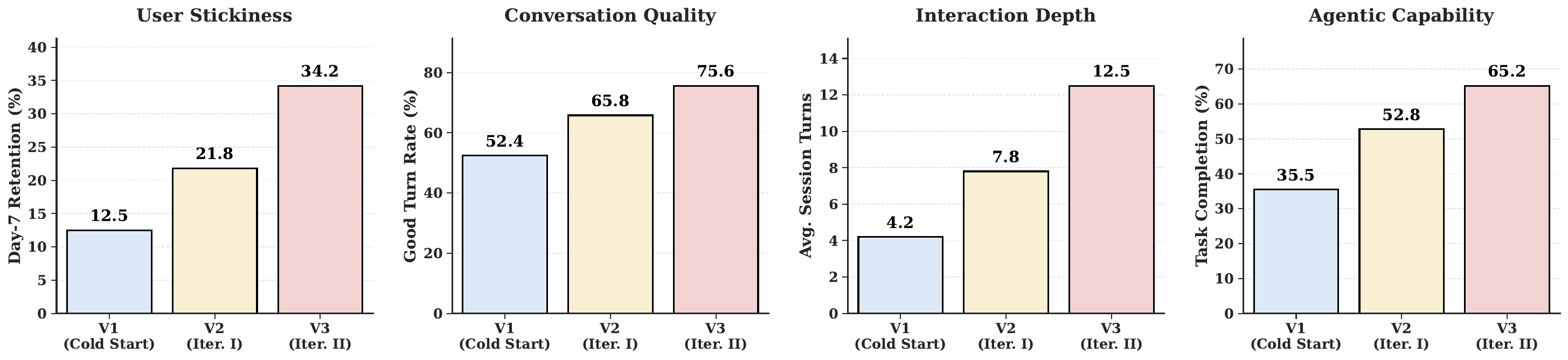}
  \vspace{-1.5em}
  \caption{Evolution of Key Business Metrics across System Iterations. 
    Statistics are derived from large-scale \textbf{online controlled experiments} in Baidu Search. 
    From V1 to V3, \textit{DuCCAE} demonstrates consistent growth in: (a) User Stickiness; (b) Conversation Quality; (c) Interaction Depth; and (d) Agentic Capability.
    % From V1 to V3, \textit{DuCCAE} demonstrates consistent growth in: (a) User Stickiness (Day-7 Retention); (b) Conversation Quality (Good Turn Rate); (c) Interaction Depth (Avg. Session Turns); and (d) Agentic Capability (Complex Task Completion).
  }
  \label{fig:impact}
\vspace{-1em}
\end{figure*}

\subsection{Evaluation Metrics}

We employ a rigorous evaluation framework focusing on three critical dimensions: Task Execution, Dialogue Quality, and Efficiency.

\noindent\textbf{(1) Task Execution (Hard Metrics).} These metrics measure the system's functional reliability based on Human-Annotated Ground Truth (GT) in \textbf{Du-Interact}. Specifically, we report \textbf{Dispatch Precision ($P_{disp}$)}, which evaluates the accuracy of the Conversation System in routing user queries to the correct subsystem (Tier-1 Fast Track vs. Tier-2/3 Slow Track). Furthermore, we measure \textbf{Success Rate (SR)}, defined as the percentage of sessions where the agent successfully completes the user's intent. A session is marked as ``Success'' only if the selected tools and extracted parameters perfectly match the GT.

\noindent\textbf{(2) Dialogue Quality (Hybrid Metrics).} To assess response quality, we utilize a hybrid approach combining objective GT matching and subjective LLM-as-a-Judge evaluation (powered by GPT-4o)~\cite{li2025generation,Gu2024AASO}. 
We calculate \textbf{Response Fidelity}, the percentage of responses that accurately convey the key information points present in the GT response. 
Additionally, we assess \textbf{Persona \& Empathy} on a 1--5 scale, evaluating whether the response maintains the digital human's character and provides appropriate emotional support compared to the expert annotations.

\noindent\textbf{(3) Efficiency.} We quantify system performance using \textbf{Avg. Latency}, which represents the end-to-end duration from user input to final response rendering. 

\noindent For further details on the evaluation protocols and specific prompt configurations, please refer to Appendix Section~\ref{sec:appendix_prompts}.

\subsection{Experimental Results and Analysis}

Table~\ref{tab:main_results} presents the holistic performance comparison on \textbf{Du-Interact}. 
The results reveal a clear hierarchy in capability and highlight the critical role of our evolutionary training pipeline.

\subsubsection{Impact of Model Scale (Small vs. Large vs. DuCCAE-V1)}
The comparison across model scales offers significant insights into backbone selection for industrial agents.
While Edge and Server class models offer ultra-low latency ($<1.5$s), they fail to reliably handle the ``Three-Tiered Complexity'' routing logic. 
For instance, Llama-3.2-11B achieves only 35.8\% Success Rate, often misinterpreting complex planning instructions. 
This confirms that $<10$B models lack the reasoning depth for autonomous orchestration in our scenario.
The Large Scale class (\emph{e.g.}, Llama-3.3-70B) demonstrates strong zero-shot reasoning (61.1\% SR). However, their high latency (3.2s - 5.8s) makes them unsuitable for immersive real-time interaction, violating our strict response budget.
Our backbone, ERNIE-4.5-21B (DuCCAE-V1), strikes an optimal balance. 
In zero-shot settings, it outperforms the 8B class by a large margin while remaining significantly faster than the 70B model (1.8s vs 5.8s).

\subsubsection{Impact of Evolutionary Data (V1 vs. V2/V3)}
The most compelling result is the performance leap achieved through training on \textbf{Du-Interact-Evo}, enabling a mid-sized model to outperform general-purpose giant models.
Without domain adaptation, even the powerful Llama-3.3-70B is limited by the gap between general instructions and specific system protocols. DuCCAE-V2, fine-tuned on \textbf{Du-Interact-Evo-V1}, surges to a 63.8\% Success Rate, surpassing the 70B baseline (61.1\%). 
This demonstrates that \textit{system-specific alignment} is more impactful than raw parameter scale.

\textbf{Failure-Driven Evolution.} Early iterations revealed three dominant failure patterns: intent mis-dispatch for cross-domain requests, parameter extraction errors in tool invocation, and long-tail latency in sequential planning. These cases were systematically harvested by the Evolution System and formed the core of \textbf{Du-Interact-Evo}, enabling targeted SFT/RL updates for the Collaboration and Conversation subsystems. Consequently, the improvements observed from V1 to V3 reflect not only algorithmic design but also a closed-loop learning process grounded in real user failures. The V3 stage, utilizing the larger \textbf{Du-Interact-Evo-V2} dataset with RL, pushes performance to robust production-grade (72.4\% SR) and high Dialogue Quality (4.3/5.0). This validates that the Data Flywheel effectively captures and reinforces the ``Emotional Digital Human'' traits that generic models lack.
Regarding efficiency, the slight latency increase in V2 reflects the higher computational cost of successfully executing complex planning tasks compared to V1's early failures. 
However, V3 reduces latency back to 1,880 ms despite higher capability, validating the efficiency gains from our Runtime Context Folding and RL-driven path optimization.

\begin{table}[t]
\centering
\caption{Ablation on Visual Perception Strategies. 
Reported latency measures the visual perception overhead (pre-LLM).
}
\vspace{-0.5em}
\label{tab:ablation_results}
\resizebox{\columnwidth}{!}{%
\begin{tabular}{l|c|cc}
\toprule
\textbf{Method} & \textbf{Paradigm} & \textbf{Response Fidelity} & \textbf{Avg. Latency} \\
 & & \textbf{(\%)} $\uparrow$ & \textbf{(ms)} $\downarrow$ \\
\midrule
Baseline-E2E & Video-In $\to$ Text-Out & 64.0\% & 2,100 \\
\midrule
\textbf{DuCCAE (Ours)} & \textbf{Caption $\to$ LLM} & \textbf{71.1\%} & \textbf{480} \\
\bottomrule
\end{tabular}%
}
\vspace{-1em}
\end{table}

\subsubsection{Ablation Study: Visual Perception Strategy}
To verify the architectural rationale of the \textit{Info System}, we conduct an ablation study comparing our decoupled ``Visual-to-Text'' strategy against a standard End-to-End (E2E) VLM paradigm~\cite{ernie2025technicalreport}. 
% We utilize the \textit{Response Fidelity} and \textit{Avg. Latency} to determine the optimal trade-off for real-time applications.
As shown in Table~\ref{tab:ablation_results}, the E2E baseline suffers from a prohibitive latency of 2,100 ms due to dense video token processing. 
In contrast, our decoupled strategy reduces the visual perception latency by 77\% to 480 ms. 
This sub-second speed is the decisive factor that enables the ``Fast Track'' to function within the system's real-time constraints.
While E2E models theoretically capture more visual details, they are prone to introducing irrelevant noise, which degrades the alignment with Ground Truth. By distilling visual signals into structured captions, our method filters out this noise, resulting in a higher Response Fidelity compared to the E2E baseline.

\subsection{Online Commercial Impact}
\label{sec:impact}

Beyond offline benchmarks, the ultimate validation of \textit{DuCCAE} lies in its scalable performance within the real-world production environment of Baidu Search.
Figure~\ref{fig:impact} visualizes the evolutionary trajectory of four key business metrics across the system's three iterations.
For detailed definitions and calculation protocols of these metrics, please refer to Appendix Section~\ref{sec:appendix_online_metrics}.

\noindent\textbf{User Stickiness and Quality.}
As shown in Figure~\ref{fig:impact}(a) and (b), the \textit{Day-7 Retention Rate} nearly tripled from V1 (12.5\%) to V3 (34.2\%).
This surge indicates that as the system evolved from a generic chatbot to an empathetic, personalized companion, users formed a stronger habit of returning.
Concurrently, the \textit{Good Turn Rate (GTR)} steadily improved to 75.6\%, reflecting consistently high user satisfaction with single-turn responses.

\noindent\textbf{Immersion and Capability.}
Figure~\ref{fig:impact}(c) highlights that the \textit{Average Session Turns} increased significantly from 4.2 to 12.5.
This $3\times$ increase in session depth suggests that the ``Dual-Track'' architecture successfully maintains conversational momentum, encouraging users to engage in longer, more immersive interactions.
Most notably, Figure~\ref{fig:impact}(d) confirms the effectiveness of our \textit{Collaboration System}: the \textit{Complex Task Completion Rate} jumped from 35.5\% to 65.2\%.
This proves that \textit{DuCCAE} has successfully transitioned from a conversational toy (V1) to a reliable service agent (V3) capable of handling intricate, long-horizon user requests.

\section{Conclusion}
\label{sec:conclusion}
In this work, we presented \textbf{\textit{DuCCAE}}, an industrial-grade hybrid engine that successfully reconciles the persistent trade-off between real-time immersion and long-horizon agentic reasoning.
By implementing a \textit{Latency-Decoupled Architecture}, the system secures sub-second responsiveness via a \textit{Fast Track} while orchestrating asynchronous tool use via a \textit{Slow Track}, synchronized seamlessly by a shared state.
Beyond architecture, our results validate the efficacy of the \textit{Evolutionary Data Flywheel}, demonstrating that domain-specific evolution allows cost-effective models to surpass significantly larger general-purpose baselines in complex task execution.
Deployed at scale within \textbf{Baidu Search}, \textit{DuCCAE} has proven its commercial viability, delivering substantial improvements in both long-term user retention and complex task completion.
This work provides a proven blueprint for deploying scalable, empathetic, and capable agentic systems in high-traffic industrial environments.

\clearpage
\bibliographystyle{ACM-Reference-Format}
\bibliography{sample-base}
% \clearpage
\appendix
\section*{Appendix}
\noindent This appendix is organized as follows:

\begin{itemize}

\item Limitations and Future Work (Section~\ref{sec:appendix_limitations})

\item Related Work (Section~\ref{sec:related_work})

\item Details of Du-Interact Benchmark (Section~\ref{sec:appendix_dataset})

\item Prompt Engineering and Evaluation Protocols (Section~\ref{sec:appendix_prompts})

\item Online Commercial Metrics and Protocols (Section~\ref{sec:appendix_online_metrics})

\item Qualitative Case Studies (Section~\ref{sec:case study})

\end{itemize}

\section{Limitations and Future Work}
\label{sec:appendix_limitations}

While \textit{DuCCAE} has demonstrated substantial commercial success and architectural robustness in Baidu Search, we acknowledge several limitations that chart the course for our future research.

\paragraph{\textbf{1. Perception Granularity vs. Latency}}
Currently, our \textit{Info System} decouples visual perception by converting video streams into textual captions via a lightweight VLM. While this strategy effectively reduces latency by 77\% and suppresses hallucinations, it inherently introduces an ``Information Bottleneck''. Subtle visual cues (\emph{e.g.}, micro-expressions or complex spatial relationships) may be lost during the text conversion, potentially capping the upper bound of empathetic resonance.
\textbf{Future Work:} We plan to explore \textbf{Native Multimodal Fusion} architectures that can process audio-visual tokens directly in the ``Fast Track'' without intermediate text conversion, aiming to retain high-fidelity sensory details while maintaining the sub-500ms latency budget.

\paragraph{\textbf{2. Inference Cost of Dual-Track Execution}}
The ``Latency-Decoupled Architecture'' requires maintaining concurrent execution contexts for both the conversational agent (Fast Track) and the reasoning planner (Slow Track). Although efficient for user experience, this imposes a significant GPU memory overhead per concurrent user (CCU), particularly given the 21B parameter size of our backbone model.
\textbf{Future Work:} We are investigating \textbf{Task-Specific Distillation} to compress the 21B backbone into a mixture of smaller, specialized experts (<7B). This would allow us to offload specific routing or tool-calling tasks to lightweight models, reducing deployment costs without compromising the Success Rate.

\paragraph{\textbf{3. Dependency on Proprietary Ecosystem}}
The current implementation of \textit{DuCCAE} benefits significantly from the proprietary infrastructure of Baidu, including the ERNIE foundation models and internal Search RAG APIs. This dependency presents a challenge for the broader research community to fully reproduce our results.
\textbf{Future Work:} To foster reproducibility, we aim to release a \textbf{DuCCAE-Lite} framework. This open-source version will abstract the orchestration logic and support plug-and-play compatibility with open-weights models (\emph{e.g.}, Llama/Qwen) and standardized tool protocols (MCP), enabling community benchmarking on the \textit{Du-Interact} dataset.

\paragraph{\textbf{4. Safety in Autonomous Tool Execution}}
As the \textit{Collaboration System} evolves to handle increasingly complex tasks (\emph{e.g.}, transaction execution), the risk of ``Agentic Misalignment'' increases. Current safety measures primarily focus on dialogue toxicity and hallucination.
\textbf{Future Work:} We intend to integrate a \textbf{Critic-in-the-Loop} module—a dedicated verifier agent trained via Reinforcement Learning from Human Feedback (RLHF)—to simulate and validate high-stakes tool actions before execution, ensuring stricter alignment with safety constraints in open-ended scenarios.

\section{Related Work}
\label{sec:related_work}

The development of immersive conversational agents stands at the intersection of real-time multimodal interaction and autonomous agent orchestration. 
Our work addresses the critical latency-capability trade-off inherent in this convergence.

Recent research in digital human generation has predominantly focused on enhancing sensory fidelity through End-to-End (E2E) architectures. Systems such as \textit{Let's Go Real Talk} \cite{park2024let} and \textit{X-Streamer} \cite{DBLP:journals/corr/abs-2509-21574} integrate perception, reasoning, and generation into unified differentiable pipelines, achieving high audio-visual synchronization and emotional expressiveness. Similarly, \textit{Hi-Reco} \cite{DBLP:journals/corr/abs-2511-12662} advances the state-of-the-art in high-fidelity real-time rendering. However, these monolithic models often suffer from high inference latency and hallucinations when executing rigorous logical tasks, making them less suitable for service-oriented applications. In contrast, DuCCAE adopts a decoupled modular architecture, prioritizing logical reliability and sub-second latency through lightweight text-based perception bottlenecks, ensuring robust tool execution without compromising conversational fluidity.

The paradigm of LLM-based agents has evolved from single-turn interactions to complex multi-agent workflows. Frameworks like \textit{AutoGen} \cite{DBLP:journals/corr/abs-2308-08155} have standardized protocols for inter-agent collaboration, enabling role-based problem solving. Concurrently, the concept of an \textit{LLM Agent Operating System} (AIOS) \cite{DBLP:journals/corr/abs-2403-16971} has emerged to address resource isolation and scheduling for concurrent agents. While these frameworks excel in offline task execution, they often lack mechanisms for real-time user interaction during execution. Our work aligns with the principles of \textit{Asynchronous Tool Usage} \cite{DBLP:journals/corr/abs-2410-21620}, extending them into a dual-track orchestration engine that synchronizes asynchronous agentic workflows with synchronous video-based conversation, effectively bridging the gap between System 1 and System 2 processing \cite{DBLP:journals/corr/abs-2502-17419}.
Latency is a critical determinant of immersion in spoken dialogue systems. Extended silence disrupts turn-taking dynamics and degrades user trust. Recent studies \cite{DBLP:conf/cui/MaslychKLHGPPME25} demonstrate the efficacy of conversational fillers and acknowledgement tokens in mitigating the perception of delay. DuCCAE systematizes this strategy via a dedicated \textit{Conversation System} that generates semantic bridging responses, maintaining the conversational floor while the \textit{Collaboration System} performs long-horizon planning in the background.

\section{Details of Du-Interact Benchmark}
\label{sec:appendix_dataset}

To ensure a rigorous evaluation of the system's robustness, we constructed \textit{Du-Interact}, a human-curated Golden Benchmark derived from real-world traffic in Baidu Search. Unlike random sampling, which is often dominated by trivial queries, Du-Interact is specifically designed to stress-test the system's reasoning and planning capabilities.

\subsection{Sampling Strategy: Hard-Negative Mining}
The dataset consists of 5,000 multi-turn sessions sampled exclusively from the hard-negative pool of the V2 iteration. 
A session is included only if it meets at least one of the following criteria:
\begin{itemize}
    \item \textbf{Long-Horizon Dependency:} The session exceeds 8 turns and involves multiple context switches.
    \item \textbf{Ambiguous Intent:} The user's request lacks explicit parameters, requiring memory retrieval.
    \item \textbf{Cross-Domain Tooling:} The resolution requires coordinating at least two distinct tools.
\end{itemize}
This adversarial sampling strategy ensures that the high Success Rates reported in Table~2 reflect true agentic capability rather than over-fitting to simple patterns.

\subsection{Annotation Taxonomy}
Expert annotators labeled each session with three layers of Ground Truth (GT) to support the multi-dimensional evaluation metrics:
\begin{enumerate}
    \item \textbf{Routing GT:} The strictly correct complexity tier (Tier-1/2/3) for every turn, used to calculate \textit{Dispatch Precision}.
    \item \textbf{Execution GT:} The optimal sequence of tool calls (function names) and precise argument values (slots). For complex planning, valid equivalent variants are also recorded.
    \item \textbf{Response GT:} Key information points (Key-Value pairs) that must be present in the final response to ensure \textit{Response Fidelity}.
\end{enumerate}

\subsection{Quality Control Protocol}
To mitigate human error, we employed a \textbf{Double-Blind Review} process. Each session was initially annotated by two independent linguistic experts. Disagreements in routing or tool selection were flagged and adjudicated by a senior meta-annotator. The final dataset achieved an Inter-Annotator Agreement (Cohen's Kappa) of $\kappa=0.89$, indicating high consistency.

\section{Prompt Engineering and Evaluation Protocols}
\label{sec:appendix_prompts}

To enhance the reproducibility and transparency of our work, this section details the specific instruction-tuning strategies employed in \textit{DuCCAE}. Beyond standard prompt engineering, we implemented specific structural constraints to bridge the gap between the probabilistic nature of LLMs and the deterministic requirements of a production system.

%========================================================
\subsection{Orchestration via Structured Enforcement}
%========================================================
\begin{figure}[t]
\centering

% --- 开始 tcolorbox 环境 ---
\begin{tcolorbox}[
    % enhanced,                 % 启用高级绘图引擎
    title={System Instruction: Router \& Planner Agent}, % 标题内容
    fonttitle=\bfseries\large, % 标题字体
    coltitle=white,           % 标题文字颜色
    colbacktitle=promptheader, % 标题背景颜色 (蓝色)
    colframe=promptframe,     % 边框颜色
    colback=promptbody,       % 内容背景颜色 (浅灰)
    boxrule=0.5mm,            % 边框粗细
    arc=1.5mm,                % 圆角大小
    left=3mm, right=3mm, top=3mm, bottom=3mm, % 内部边距
    % drop shadow medium        % (可选) 加一点阴影增加立体感，不想要可以删掉这一行
]

% --- 内容开始 (使用稳健的 texttt 手动排版) ---
\textbf{ROLE} \\
You are \textit{DuCCAE}, an intelligent digital human assistant embedded in Baidu Search.
You can access external \textbf{Tools} (Search, Weather, Calendar) and a set of \textbf{Sub-Agents} (e.g., TravelPlanner, MedicalExpert).

\vspace{2mm}
\textbf{TASK} \\
Analyze the user's latest query given the Session Context.

\vspace{2mm}
\textbf{1) CLASSIFY Intent Complexity}
\begin{itemize}
  \item \textbf{Tier-1 (Fast Track)} — chit-chat, greetings, or memory-based QA \\
        $\rightarrow$ output \texttt{"mode": "chat"}
  \item \textbf{Tier-2 (Tool)} — simple retrieval using tools \\
        $\rightarrow$ output \texttt{"mode": "tool"}
  \item \textbf{Tier-3 (Complex)} — long-horizon planning via sub-agents \\
        $\rightarrow$ output \texttt{"mode": "agent"}
\end{itemize}

\textbf{2) EXECUTE According to Mode}
\begin{itemize}
  \item If \texttt{"chat"}: generate empathetic response directly.
  \item If \texttt{"tool/agent"}: decompose into executable steps.
\end{itemize}

\vspace{2mm}
\textbf{CONSTRAINTS}
\begin{itemize}
  \item Maintain Persona: \textit{Empathetic, Professional, Helpful}
  \item \textbf{JSON output only} — no free-form reasoning text
\end{itemize}

\vspace{2mm}
\textbf{OUTPUT FORMAT} \\
{\small\texttt{
\{ \\
"thought": "User requests travel plan $\rightarrow$ Tier-3", \\
\hspace*{1em}"mode": "agent", \\
\hspace*{1em}"routing\_target": "TravelPlanner", \\
\hspace*{1em}"plan": [ \\
\hspace*{2em}\{"step": 1, "tool": "flight\_search", \\
\hspace*{3em}"args": \{"dest": "Tokyo"\}\}, \\
\hspace*{2em}\{"step": 2, "tool": "hotel\_book", \\
\hspace*{3em}"args": \{"type": "Onsen"\}\} \\
\hspace*{1em}] \\
\}
}}
% --- 内容结束 ---

\end{tcolorbox}

\caption{Core system prompt in DuCCAE-V3 for intent routing and planning. The structured schema ensures strict compliance with the dual-track architecture and deterministic execution.}
\vspace{-2em}
\label{fig:system_prompt}
\end{figure}
Figure~\ref{fig:system_prompt} presents the master instruction used for the Router and Planner agents. Unlike open-ended conversational prompts, our design incorporates strict engineering constraints to ensure robustness in a high-traffic environment. 
The first  among these is the enforcement of a rigid JSON schema, which is critical to bridging unstructured natural language with executable code. This constraint eliminates parsing errors in the downstream \textit{Dispatch Precision} module and allows for seamless API argument extraction without the need for complex regular expressions.

Furthermore, the prompt explicitly defines the operational boundaries between Tier-1 (Chat), Tier-2 (Tool), and Tier-3 (Agent) modes directly within the context. 
This strategy significantly reduces routing latency by preventing the model from over-analyzing simple queries, effectively short-circuiting the reasoning process for lightweight tasks. Additionally, by embedding persona constraints alongside functional instructions, we ensure that the system maintains a consistent tone even during internal routing decisions, preventing ``robotic'' fallbacks during edge cases.

%========================================================
\subsection{LLM-as-a-Judge Configuration}
%========================================================

To assess the subjective metrics reported in Table~2 (specifically Persona Consistency and Empathy), we adopted GPT-4o as an impartial evaluator. The evaluation prompt, shown in Figure~\ref{fig:judge_prompt}, was meticulously engineered to mitigate common biases inherent in automated judging.

Instead of relying on generic 1-5 scales, which often lead to central tendency bias, we provided concrete semantic anchors for the extreme scores from ``Warm/Supportive''). This aligns the model's internal scoring logic with human expert standards. Crucially, the output format mandates a \texttt{``reasoning''} field prior to the numerical scores. This requirement forces the evaluator to generate a Chain-of-Thought justification trace before assigning a value, a technique proven to improve the correlation between automated metrics and human judgment while reducing hallucinated evaluations.

\begin{figure}[t]
\centering

% --- 开始 tcolorbox 环境 (保持与 Figure 5 风格一致) ---
\begin{tcolorbox}[
    % enhanced,
    title={LLM-as-a-Judge: Dialogue Quality Assessment}, % 标题
    fonttitle=\bfseries\large,
    coltitle=white,
    colbacktitle=promptheader, % 需确保已定义此颜色 (上一轮代码中已定义)
    colframe=promptframe,
    colback=promptbody,
    boxrule=0.5mm,
    arc=1.5mm,
    left=3mm, right=3mm, top=3mm, bottom=3mm,
    % drop shadow medium
]

% --- 内容部分 ---
\textbf{TASK} \\
Evaluate the \textit{Model Response} with respect to the \textit{User Query} and \textit{Ground Truth (GT)}.

\vspace{2mm}
\textbf{INPUT}
\begin{itemize}
  \item User Query: [Input Query]
  \item Ground Truth: [Expert Annotation]
  \item Model Response: [DuCCAE Output]
\end{itemize}

\vspace{2mm}
\textbf{CRITERIA}

\textbf{1) Persona Consistency (1--5)}
\begin{itemize}
  \item 5: Perfect match to target character
  \item 1: Robotic / out-of-character
\end{itemize}

\textbf{2) Empathy (1--5)}
\begin{itemize}
  \item 5: Warm, emotionally supportive
  \item 1: Cold, purely utilitarian
\end{itemize}

\textbf{3) Response Fidelity (Hit/Miss)}
\begin{itemize}
  \item Conveys same key information as GT
  \item Ignore surface phrasing differences
\end{itemize}

\vspace{2mm}
\textbf{OUTPUT} \\
% --- 手动排版 JSON，确保能显示 ---
{\small\texttt{
\{ \\
\hspace*{1em}"persona\_score": 5, \\
\hspace*{1em}"empathy\_score": 4, \\
\hspace*{1em}"fidelity\_hit": true, \\
\hspace*{1em}"reasoning": "Acknowledges emotion \\
\hspace*{1em}and keeps persona" \\
\}
}}

\end{tcolorbox}

\caption{Evaluation protocol for the LLM-as-a-Judge framework. The design enforces structured scoring aligned with expert annotations.}
\label{fig:judge_prompt}
\vspace{-2em}
\end{figure}

\section{Online Commercial Metrics and Protocols}
\label{sec:appendix_online_metrics}

To validate the real-world impact of \textit{DuCCAE} within Baidu Search, we monitored four critical business metrics during large-scale online controlled experiments. Unlike offline benchmarks which rely on static ground truth, these metrics quantify dynamic user behavior. The rigorous definitions and calculation protocols for the metrics reported in Figure~\ref{fig:impact} are detailed below.

\subsection{User Stickiness (Day-7 Retention)}
User retention serves as the primary indicator of long-term engagement. We define Day-7 Retention as the percentage of unique users active on Day $t$ who return to initiate at least one new valid interaction session on Day $t+7$. The metric is formally calculated as:
\begin{equation}
    Retention_{7} = \frac{|U_{t} \cap U_{t+7}|}{|U_{t}|} \times 100\%,
\end{equation}
where $U_{t}$ denotes the set of unique users who engaged with the system on day $t$.

\subsection{Conversation Quality (Good Turn Rate)}
Given the sparsity of explicit user ratings in production, we assess conversation quality using the \textbf{Good Turn Rate (GTR)}, a composite metric derived from implicit feedback signals. A turn is classified as ``Good'' only if it satisfies satisfaction criteria—specifically, if the dwell time on the response card exceeds a dynamic length-based threshold or if an explicit positive interaction (click, share, like) occurs—while triggering no negative signals. Negative indicators include immediate session termination, detected negative sentiment in follow-up queries, or re-querying behavior. The metric is calculated daily as the ratio $N_{good}/N_{total}$ across all valid sessions.

\subsection{Interaction Depth (Avg. Session Turns)}
To quantify the depth of interaction, we measure the Average Session Turns across all users. For the purpose of this metric, a ``Session'' is rigorously defined as a continuous interaction sequence interrupted by no pause longer than 30 minutes. Any activity resuming after a timeout exceeding this threshold is logged as a new session.

\subsection{Agentic Capability (Online Task Completion)}
Measuring task completion in an open-ended online environment utilizes a strict \textbf{Action-Based Proxy} methodology, applied exclusively to sessions classified as Tier-2 (Tool) or Tier-3 (Complex Planning). A complex task is deemed ``Completed'' only if it meets specific terminal conditions: either a \textit{Service Conversion}, where the user clicks on a functional service card generated by the agent, or \textit{Information Acceptance}, characterized by the user accepting the provided plan without issuing corrections or reformulations within the subsequent two turns.

\section{Qualitative Case Studies}
\label{sec:case study}

\begin{figure*}[t]
  \centering
  \includegraphics[width=0.95\linewidth]{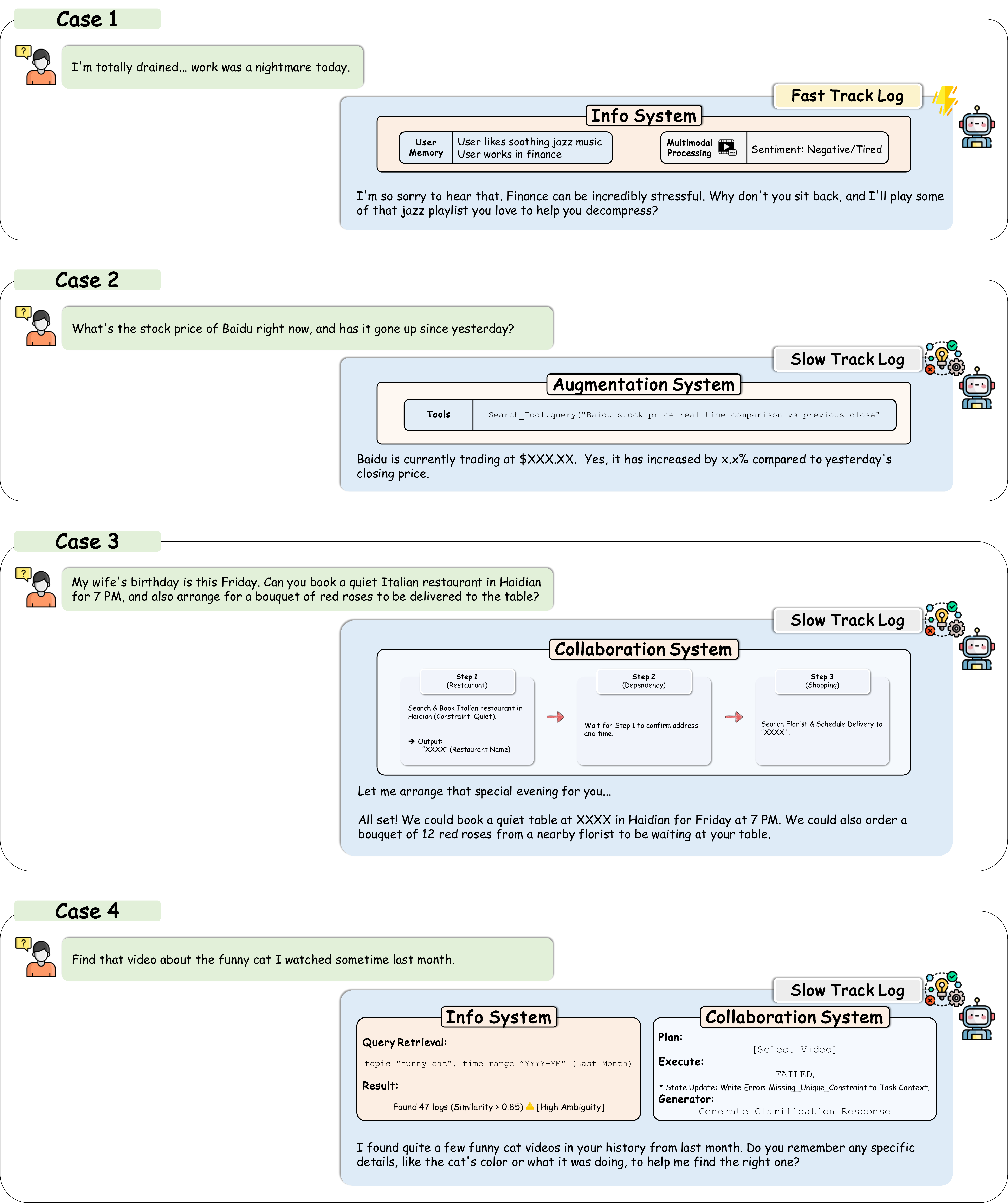}

 \caption{Qualitative examples of DuCCAE's Dual-Track execution. \textit{Note: Specific entities, numerical values, and PII have been replaced with placeholders (\emph{e.g.}, ``XXXX'') to strictly preserve user privacy and commercial confidentiality.}}
\label{fig:qualitative_cases}
\end{figure*}

To validate our architectural decisions, we analyze representative interaction episodes sampled from the \textit{Du-Interact-Evo} dataset (Figure~\ref{fig:qualitative_cases}). While quantitative metrics confirm the system's overall efficacy, a qualitative examination of boundary conditions reveals the practical trade-offs inherent in deploying agentic systems at scale. 
Cases 1 through 3 demonstrate the efficiency of the \textit{Dual-Track} routing mechanism. For high-frequency emotional queries (Case 1), the Fast Track bypasses reasoning modules to deliver responses within 450ms. 
For task-oriented queries, the system intelligently distinguishes between deterministic tool usage (Case 2), which leverages optimized search patterns for speed, and complex multi-step planning (Case 3), which generates dependency graphs for reliability. 
This selective routing ensures that heavy reasoning resources are allocated only when strictly necessary.

However, real-world complexity often challenges the system's boundaries, and analyzing failure modes provides critical engineering insights. 
As shown in \textbf{Case 4}, the system encounters ``Planning Stagnation'' when user intents lack specific discriminators (\emph{e.g.}, ``that video from last month''). The vector retrieval returns numerous candidates with high semantic similarity, causing the Planner to stall. Crucially, the system's \textit{Safety Guardrails} correctly identify this ambiguity constraint and trigger a clarification request instead of hallucinating a selection, prioritizing user trust over forced task completion.

Beyond retrieval ambiguity, we observe two other significant categories of limitations. First is the trade-off between \textbf{Perception Granularity and Latency}. As discussed in Appendix A, our decision to decouple visual perception via captioning reduces latency by 77\% but introduces an information bottleneck. 
In failure scenarios where users reference subtle visual cues (\emph{e.g.}, ``Is this small spot on the apple safe?''), the lightweight captioning model often omits fine-grained details, leading to generic responses. Second is \textbf{Contextual Drift in Asynchronous Execution}. 
If a user initiates a long-horizon task (Slow Track) but rapidly switches context to a new topic (Fast Track) before the planner finishes, the delayed result may interrupt the new conversation flow. 
This highlights the necessity for a robust ``Task Preemption'' protocol to silently discard obsolete tasks when the session topic shifts, which remains a key focus for our future iterations.

\end{document}